\documentclass[11pt]{article}
\usepackage{hyperref}
\usepackage{acl2015}
\usepackage{times}

\usepackage{amsmath}
\usepackage{amssymb}
\usepackage{amsthm}

\usepackage{booktabs}
\usepackage{tabularx}
\usepackage{tikz}
\usepackage{subcaption}
\usepackage{ragged2e}

\usepackage{algorithm}
\usepackage{algpseudocode}
\usepackage{algorithmicx}

\usepackage{siunitx}
\usepackage{arydshln}

\usepackage[bb=fourier]{mathalfa}
\usepackage[font=small]{caption}

\pdfpagewidth=\paperwidth
\pdfpageheight=\paperheight

\newcommand{\reals}{\mathbb{R}}
\newcommand{\labels}{\mathcal{L}}

\newcommand{\texts}{\mathbf{x}}
\newcommand{\paths}{\mathbf{y}}
\newcommand{\txt}{x}
\newcommand{\pth}{y}
\newcommand{\aligns}{\mathbf{a}}

\newcommand{\indicate}{\mathbf{1}}

\newenvironment{ditemize}
{\begin{itemize}\setlength{\itemsep}{0pt}}
{\end{itemize}}

\newcommand{\whdashline}{\\[-0.8em]\hdashline\\[-0.8em]}

\title{Alignment-Based Compositional Semantics for Instruction Following}

\author{Jacob Andreas \and Dan Klein \\
  Computer Science Division \\
  University of California, Berkeley \\
  {\tt \{jda,klein\}@cs.berkeley.edu}}

\date{}

\begin{document}

\maketitle

\begin{abstract}
  This paper describes an alignment-based model for interpreting natural
  language instructions in context.  We approach instruction following as a
  search over plans, scoring sequences of actions conditioned on structured
  observations of text and the environment. By explicitly modeling
  both the low-level compositional structure of individual actions and the
  high-level structure of full plans, we are able to learn both grounded
  representations of sentence meaning and pragmatic constraints on
  interpretation. To demonstrate the model's flexibility, we apply it to a
  diverse set of benchmark tasks. On every task, we outperform strong
  task-specific baselines, and achieve several new state-of-the-art results.
\end{abstract}

\section{Introduction}

In instruction-following tasks, an agent executes a sequence of actions
in a real or simulated environment, in response to a sequence of natural
language commands. Examples include giving navigational directions to robots
and providing hints to automated game-playing agents. 
Plans specified with natural language 
exhibit compositionality both at the level of individual actions and at the
overall sequence level. This paper describes a framework for learning to follow
instructions by leveraging structure at both levels. 

%

Our primary contribution is a new, alignment-based approach to grounded
compositional semantics. Building on related logical approaches
\cite{Reddy14Freebase,Pourdamghani14AMRAlign}, we recast instruction following
as a pair of nested, structured alignment problems. Given instructions and a
candidate plan, the model infers a \emph{sequence-to-sequence} alignment between
sentences and atomic actions. Within each sentence--action pair, the
model infers a \emph{structure-to-structure} alignment between the syntax of the
sentence and a graph-based representation of the action.

\begin{figure*}
  \setlength{\tabcolsep}{11pt}
  \begin{tabularx}{\textwidth}{XXX}
    \centering
    \includegraphics[height=1.1in, clip=true, trim=0in .3in 7.4in 6in]{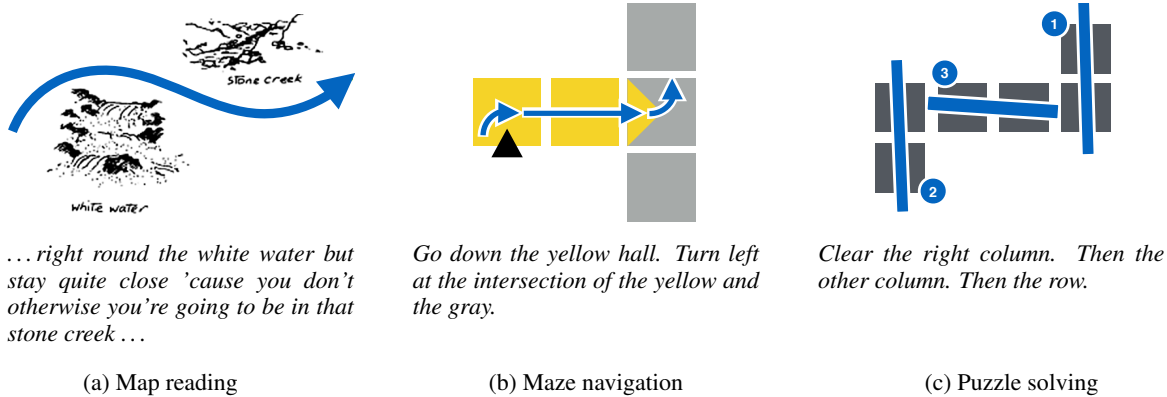}
    &
    \centering
    \includegraphics[height=1.2in, clip=true, trim=0in 4.8in 7.4in 0in]{examples}
    &
    \centering\arraybackslash
    \includegraphics[height=1.2in, clip=true, trim=7.4in 5.0in 0in 0in]{examples} 
    \\
    \footnotesize\it \ldots right round the white water but stay quite close 'cause you
    don't otherwise you're going to be in that stone creek \ldots
    &
    \footnotesize\it Go down the yellow hall. Turn left at the intersection of the
    yellow and the gray.
    &
    \footnotesize\it Clear the right column. Then the other column. Then the
    row.
  \end{tabularx}
  \vspace{.5em}

  \begin{subfigure}[b]{0.3\textwidth}
    \caption{Map reading}
    \label{fig:tasks:hcrc}
  \end{subfigure}\hfill
  \begin{subfigure}[b]{0.3\textwidth}
    \caption{Maze navigation}
    \label{fig:tasks:sail}
  \end{subfigure}\hfill
  \begin{subfigure}[b]{0.3\textwidth}
    \caption{Puzzle solving}
    \label{fig:tasks:crossblock}
  \end{subfigure}
  \caption{Example tasks handled by our framework. The tasks feature
  noisy text, over- and under-specification of plans, and challenging search
  problems.}
  \label{fig:tasks}
\end{figure*}
At a high level, our agent is a block-structured, graph-valued conditional
random field, with alignment potentials to relate instructions to actions and
transition potentials to encode the environment model (\autoref{fig:overview}).
Explicitly modeling sequence-to-sequence alignments between text and actions
allows flexible reasoning about action sequences, enabling the agent to
determine which actions are specified (perhaps redundantly) by text, and which
actions must be performed automatically (in order to satisfy pragmatic
constraints on interpretation). 
Treating instruction following as a sequence prediction
problem, rather than a series of independent decisions
\cite{Branavan09PG,Artzi13Navigation}, makes it possible to use general-purpose
planning machinery, greatly increasing inferential power.


The fragment of semantics necessary
to complete most instruction-following tasks is essentially predicate--argument
structure, with limited influence from quantification and scoping. Thus
the problem of sentence interpretation can reasonably be modeled as one of
finding an alignment between language and the environment it describes. We allow
this structure-to-structure alignment---an ``overlay'' of language onto the
world---to be mediated by linguistic structure (in the form of dependency
parses) and structured perception (in what we term \emph{grounding graphs}).
Our model thereby reasons directly about the relationship between language and
observations of the environment, without the need for an intermediate logical
representation of sentence meaning. This, in turn, makes it possible to
incorporate flexible feature representations 
that have been difficult to integrate with previous work in semantic parsing.


We apply our approach to three established instruction-following benchmarks: the
map reading task of \newcite{Vogel10SARSA}, the maze navigation task of
\newcite{MacMahon06SAIL}, and the puzzle solving task of \newcite{Branavan09PG}.
An example from each is shown in \autoref{fig:tasks}. These benchmarks exhibit a
range of qualitative properties---both in the length and complexity of their
plans, and in the quantity and quality of accompanying language. Each task has
been studied in isolation, but we are unaware of any published approaches
capable of robustly handling all three. Our general model outperforms strong,
task-specific baselines in each case, achieving 
relative error reductions of 15--20\% over 
several state-of-the-art results. Experiments demonstrate the importance of our
contributions in both compositional semantics and search over plans. We have
released all code for this project at \linebreak
\scalebox{0.91}{\url{github.com/jacobandreas/instructions}}.

\section{Related work}

Existing work on instruction following can be roughly divided into two families:
semantic parsers and linear policy estimators.  

\paragraph{Semantic parsers} Parser-based approaches
\cite{Chen11Navigation,Artzi13Navigation,Kim13Reranker,Tellex11Commands} map
from text into a formal language representing commands. These take familiar
structured prediction models for semantic parsing
\cite{Zettlemoyer05CCG,Wong06WASP}, and train them with task-provided
supervision. Instead of attempting to match the structure of a
manually-annotated semantic parse, semantic parsers for instruction following
are trained to maximize a reward signal provided by black-box execution of the
predicted command in the environment. (It is possible to think of response-based
learning for question answering \cite{Liang13DCS} as a special case.) 

This approach uses a well-studied mechanism for
compositional interpretation of language, but is subject to certain limitations.
Because the environment is manipulated only through black-box execution of the
completed semantic parse, there is no way to incorporate current or future
environment state into the scoring function.
It is also in general necessary to
hand-engineer a task-specific formal language for describing agent behavior. 
Thus it is extremely difficult to work with environments that cannot be modeled
with a fixed inventory of predicates (e.g.\ those involving novel strings or
arbitrary real quantities).

Much of contemporary work in this family is evaluated on the maze navigation
task introduced by \newcite{MacMahon06SAIL}. \newcite{Dukes13Blocks} also
introduced a ``blocks world'' task for situated parsing of spatial robot
commands.

\paragraph{Linear policy estimators}
An alternative family of approaches is based on learning a policy over primitive
actions directly \cite{Branavan09PG,Vogel10SARSA}.\footnote{This is distinct
  from semantic parsers in which greedy inference happens to have an
  interpretation as a policy \cite{Vlachos14FuckYou}.} 
  Policy-based approaches
  instantiate a Markov decision process representing the action domain, and
  apply standard supervised or reinforcement-learning approaches to learn a
  function for greedily selecting among actions. In linear policy approximators,
  natural language instructions are incorporated directly into state
  observations, and reading order becomes part of the action selection process.  

Almost all existing policy-learning approaches make use of an unstructured
parameterization, with a single (flat) feature vector representing all text and
observations. Such approaches are thus restricted to problems that are simple enough (and
have small enough action spaces) to be effectively characterized in this
fashion. While there is a great deal of flexibility in the choice of feature
function (which is free to inspect the current and future state of the
environment, the whole instruction sequence, etc.), standard linear policy
estimators have no way to model compositionality in language or actions.

Agents in this family have been evaluated on a variety of tasks, including
map reading \cite{Anderson91MapTask} and gameplay \cite{Branavan09PG}.

\paragraph{} 
Though both families address the same class of instruction-following
problems, they have been applied to a totally disjoint set of tasks.  It should
be emphasized that there is nothing inherent to policy learning that prevents
the use of compositional structure, and nothing inherent to general
compositional models that prevents more complicated dependence on environment
state. 
Indeed, previous work \cite{Branavan11Civ,Narasimhan15DeepRL} uses
aspects of both to solve a different class of gameplay problems.
In some sense, our goal in this paper is simply to combine the strengths
of semantic parsers and linear policy estimators for fully general instruction
following.
As we shall see, however, this requires changes to many aspects of
representation, learning and inference.

\section{Representations}

\begin{figure}[t]
  \centering
  \includegraphics[width=\columnwidth, clip=true, trim=2.3in 0in 2.8in 0in]{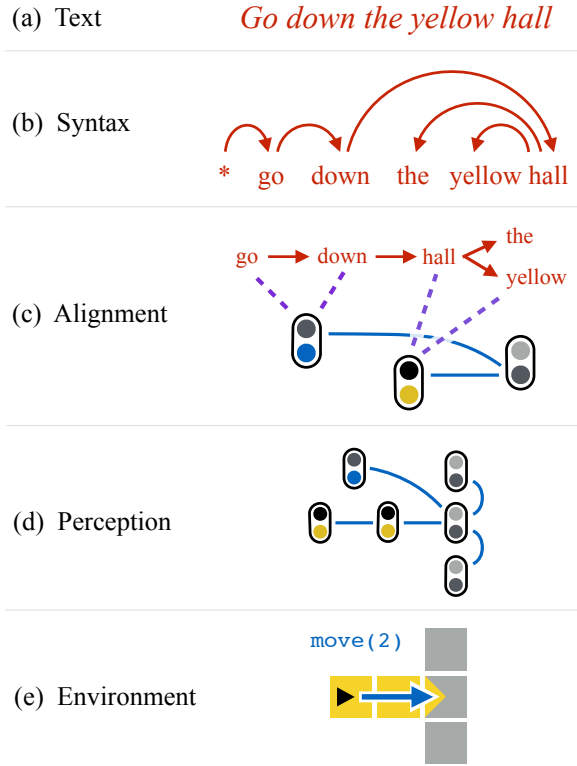}
  \caption{Structure-to-structure alignment connecting a single sentence (via
    its syntactic analysis) to the environment state (via its grounding graph).
    The connecting alignments take the place of a traditional semantic parse and
    allow flexible, feature-driven linking between lexical primitives and
    perceptual factors.}
  \label{fig:structures}
\end{figure}

We wish to train a model capable of following commands in a simulated
environment.
We do so by presenting the model with a sequence of training pairs $(\texts,
\paths)$, where each $\texts$ is a sequence of \emph{natural language
instructions} $(\txt_{1}, \txt_{2}, \dots, \txt_{m})$, e.g.:
\[ (\textit{Go down the yellow hall.},\ \ \textit{Turn left.},\ \ \dots) \]
and each $\paths$ is a demonstrated \emph{action sequence}
$(\pth_{1}, \pth_{2}, \dots, \pth_{n})$, e.g.: 
\[ (\texttt{rotate(90)},\ \ \texttt{move(2)},\ \ \dots) \]
Given a start state, $\paths$ can equivalently be characterized by a sequence of
(state, action, state) triples resulting from execution of the environment
model. An example instruction is shown in \autoref{fig:structures}a. An
example action, situated in the environment where it occurs, is shown in
\autoref{fig:structures}e. 



Our model performs compositional interpretation of instructions by leveraging
existing structure inherent in both text and actions. Thus we interpret $x_i$
and $y_j$ not as raw strings and primitive actions, but rather as structured
objects.

\paragraph{Linguistic structure}

We assume access to a pre-trained parser, and in particular that each of the
instructions $x_{i}$ is represented by a tree-structured dependency
parse. An example is shown in \autoref{fig:structures}b.


\paragraph{Action structure}




By analogy to the representation of instructions as parse trees, we assume that
each (state, action, state) triple (provided by the environment model) can be
characterized by a \emph{grounding graph}.\footnote{We note that the instruction
  following model of \newcite{Tellex11Commands} features a similarly named
  ``Generalized Grounding Graph'' (G$^3$) formalism. A G$^3$ links the syntax of
  the input command to the action ultimately executed, and is thus more
  analogous to our structured alignment variable (\autoref{fig:structures}c)
  than our perceptual representation.
}
The structure and content of this
representation is task-specific. An example grounding graph for the maze
navigation task is shown in \autoref{fig:structures}d. The example contains
a node corresponding to the primitive action \texttt{move(2)} (in the upper
left), and several nodes corresponding to locations in the environment that are
visible after the action is performed.

Each node in the graph (and, though not depicted, each edge) is decorated with a
list of features. These features might be simple indicators (e.g.\ whether the
primitive action performed was {\tt move} or {\tt rotate}), real values (the
distance traveled) or even string-valued (English-language names of visible
landmarks, if available in the environment description).
Formally, a grounding graph consists of a tuple $(V, E,
\labels, f_V, f_E)$, with
\begin{ditemize}
  \item $V$ a set of vertices
  \item $E \in V \times V$ a set of (directed) edges
  \item $\labels$ a space of labels (numbers, strings, etc.)
  \item $f_V: V \to 2^\labels$ a vertex feature function
  \item $f_E: E \to 2^\labels$ an edge feature function
\end{ditemize}

In this paper we have tried to remain agnostic to details of graph construction.
Our goal with the grounding graph framework is simply to accommodate a wider
range of modeling decisions than allowed by existing formalisms. Graphs might be
constructed directly, given access to a structured virtual environment (as in
all experiments in this paper), or alternatively from outputs of a perceptual
system. For our experiments, we have remained as close as possible to
task representations described in the existing literature. Details for each task
can be found in the accompanying software package.

Graph-based representations are extremely common in formal semantics
\cite{Jones12HERG,Reddy14Freebase}, and the version presented here
corresponds to a simple generalization of familiar formal methods. Indeed, if
$\labels$ is the set of all atomic entities and relations, $f_V$ returns a
unique label for every $v \in V$, and $f_E$ always returns a vector with one
active feature, we recover the existentially-quantified portion of first order
logic exactly, and in this form can implement large parts of classical
neo-Davidsonian semantics \cite{Parsons90NeoDavidsonian} using grounding graphs.

Crucially, with an appropriate choice of $\labels$ this formalism also makes it
possible to go beyond set-theoretic relations, and incorporate string-valued
features (like names of entities and landmarks) and real-valued features (like
colors and positions) as well.


\paragraph{Lexical semantics}

We must eventually combine features provided by parse trees with features
provided by the environment. Examples here might include simple conjunctions
({\tt word=yellow $\land$ rgb=(0.5, 0.5, 0.0)}) or more complicated computations like
edit distance between landmark names and lexical items.  Features of the
latter kind make it possible to behave correctly in environments containing
novel strings or other features unseen during training.  

This aspect of the syntax--semantics interface has been troublesome for some
logic-based approaches:~while past work has used related machinery for selecting
lexicon entries \cite{Berant14Paraphrasing} or for rewriting logical forms
\cite{Kwiatkowski2013Matching}, the relationship between text and the
environment has ultimately been mediated by a discrete (and indeed finite)
inventory of predicates.
Several recent papers have investigated simple grounded models with real-valued
output spaces \cite{Andreas14Paths,McMahan15Colors}, but we are unaware of any
fully compositional system in recent literature that can incorporate
observations of these kinds.

Formally, we assume access to a joining feature function $\phi : (2^\labels \times
2^\labels) \to \reals^d$. As with grounding graphs, our goal is to make the
general framework as flexible as possible, and for individual experiments have
chosen $\phi$ to emulate modeling decisions from previous work.

\section{Model}

\begin{figure}
\centering
\includegraphics[width=\columnwidth, clip=true, trim=0in 0in 0in 2in]{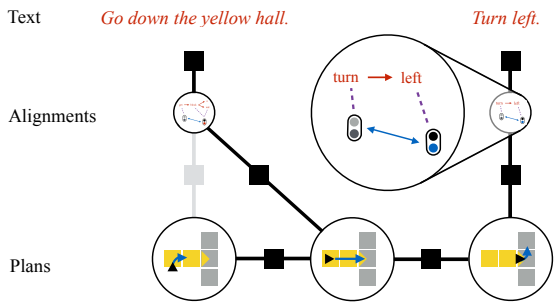}
\caption{Our model is a conditional random field that describes distributions
over state-action sequences conditioned on input text. Each variable's domain
is a structured value. Sentences align to a subset of the state--action
sequences, with the rest of the states filled in by pragmatic (planning)
implication.  State-to-state structure represents planning constraints
(environment model) while state-to-text structure represents compositional
alignment.  All potentials are log-linear
and feature-driven.}
\label{fig:overview}
\end{figure}

As noted in the introduction, we approach instruction following as a sequence
prediction problem. Thus we must place a distribution over sequences of actions
conditioned on instructions.
We decompose the problem into two components, describing interlocking models of
``path structure'' and ``action structure''.  Path structure captures how
sequences of instructions give rise to sequences of actions, while action
structure captures the compositional relationship between individual utterances
and the actions they specify. 

\subsection*{Path structure: aligning utterances to actions}

The high-level path structure in the model is depicted in 
\autoref{fig:overview}.
Our goal here is to permit both under- and over-specification of plans, and to
expose a planning framework which allows plans to be computed with lookahead
(i.e.\ non-greedily).



These goals are achieved by introducing a sequence of latent alignments between
instructions and actions.  
Consider the multi-step example in \autoref{fig:tasks:sail}. If the first
instruction {\it go down the yellow hall} were interpreted immediately, we would
have a presupposition failure---the agent is facing a wall, and cannot move
forward at all. Thus an implicit {\tt rotate} action, unspecified by text, must
be performed before any explicit instructions can be followed.

To model this, we take the
probability of a (text, plan, alignment) triple to be log-proportional
to the sum of two quantities:
\begin{enumerate}
  \item a path-only score $\psi(n;\theta) + \sum_j \psi(y_j;\theta)$
  \item a path-and-text score, itself the sum of all pair scores $\psi(x_i,
    y_j ; \theta)$ licensed by the alignment
\end{enumerate}
(1) captures our desire for pragmatic constraints on interpretation, and
provides a means of encoding the inherent plausibility of paths.
We take $\psi(n;\theta)$ and $\psi(y;\theta)$ to be linear functions of
$\theta$. (2) provides context-dependent interpretation of text by means of the
structured scoring function $\psi(\txt, \pth; \theta)$, described in the next
section. 

Formally, we associate with each instruction $x_{i}$ a sequence-to-sequence
alignment variable $a_{i} \in 1\ldots n$ (recalling that $n$ is the number of
actions).  Then we have\footnote{Here and in the remainder of this paper, we suppress the dependence of the
various potentials on $\theta$ in the interest of readability.}
\begin{align}
  p(\paths, &\aligns | \texts; \theta) \propto 
  \exp\bigg\{ \psi(n) \nonumber 
  + \sum_{j=1}^n \psi(\pth_j) \nonumber \\
  &\qquad + \sum_{i=1}^m \sum_{j=1}^n
  \indicate[a_j = i]\ \psi(\txt_i, \pth_j)\bigg\}
  \label{eq:high-level}
\end{align}
We additionally place a monotonicity constraint on the alignment variables.
This model is globally normalized, and for a fixed alignment is equivalent to a
linear-chain CRF. In this sense it is analogous to IBM Model I
\cite{Brown93IBM}, with the structured potentials $\psi(\txt_i, \pth_j)$ taking
the place of lexical translation probabilities. While alignment models from
machine translation have previously been used to align words to fragments of
semantic parses \cite{Wong06WASP,Pourdamghani14AMRAlign}, we are unaware of such
models being used to align entire instruction sequences to demonstrations.

\subsection*{Action structure: aligning words to percepts}



Intuitively, this scoring function $\psi(\txt, \pth)$ should capture how well a
given utterance describes an action. If neither the utterances nor the actions
had structure (i.e.\ both could be represented with simple bags of features), we
would recover something analogous to the conventional policy-learning approach. As
structure is essential for some of our tasks, $\psi(\txt, \pth)$ must instead
fill the role of a semantic parser in a conventional compositional model.

Our choice of $\psi(x, y)$ is driven by the following fundamental assumptions:
\emph{Syntactic relations approximately represent semantic relations. Syntactic
proximity implies relational proximity.} In this view, there is an additional
hidden structure-to-structure alignment between the grounding graph and the parsed text describing it.
\footnote{It is formally possible to regard the sequence-to-sequence and
  structure-to-structure alignments as a single (structured) random variable.
  However, the two kinds of alignments are
  treated differently for purposes of inference, so it is useful to maintain a
  notational distinction.
} Words line up with nodes, and
dependencies line up with relations. 
Visualizations are shown in
\autoref{fig:structures}c and the zoomed-in portion of \autoref{fig:overview}.

As with the top-level alignment variables, this approach can viewed as a simple
relaxation of a familiar model. CCG-based parsers assume that syntactic type
strictly determines semantic type, and that each lexical item is associated with
a small set of functional forms. Here we simply allow all words to license all
predicates, multiple words to specify the same predicate, and some edges to be
skipped. We instead rely on a scoring function to impose soft versions of the
hard constraints typically provided by a grammar. Related models have previously
been used for question answering \cite{Reddy14Freebase,Pasupat15Parser}.

For the moment let us introduce variables $b$ to denote these
structure-to-structure alignments. (As will be seen in
the following section, it is straightforward to marginalize over all choices of
$b$.  Thus the structure-to-structure alignments are never explicitly
instantiated during inference, and do not appear in the final form of $\psi(x,
y)$.) For a fixed alignment, we define $\psi(x,y,b)$ according to
a recurrence relation. Let $\txt^i$ be the $i$th word of the sentence, and let
$\pth^j$ be the $j$th node in the action graph (under some topological
ordering). Let $c(i)$ and $c(j)$ give the indices of the dependents of $\txt^i$
and children of $\pth^j$ respectively. Finally, let $\txt^{ik}$ and $\pth^{jl}$
denote the associated dependency type or relation.  Define a ``descendant''
function:
\begin{align}
  d(i,j) = \big\{ (k, l):\ k \in c(i),\ l \in c(j),\ (k,l) \in b \big\} \nonumber
\end{align}
Then,
\begin{align}
  \psi(&\txt^i, \pth^j, b) =  \exp\bigg\{ \theta^\top \phi(\txt^i, \pth^j) \nonumber
    \\ &+ \sum_{(k,l) \in d(x,y)} \Big[ \theta^\top
      \phi\big(\txt^{ik}, \pth^{jl}\big) \nonumber
       \cdot \ \psi(\txt^k, \pth^l, b) \Big] \bigg\}
\label{eq:overlay}
\end{align}
This is just an unnormalized synchronous derivation between $\txt$ and
$\pth$---at any aligned (node, word) pair, the score for the entire derivation is
the score produced by combining that word and node, times the scores at all the
aligned descendants. Observe that as long as there are no cycles in the
dependency parse, it is perfectly acceptable for the relation graph to contain
cycles and even self-loops---the recurrence still bottoms out appropriately.

\section{Learning and inference}

Given a sequence of training pairs $(\texts, \paths)$, we wish to find a parameter
setting that maximizes $p(\paths|\texts;\theta)$. If there were no latent
alignments $a$ or $b$, this would simply involve minimization of a convex
objective.
The presence of latent variables complicates things.
Ideally, we would like to sum over the latent variables,
but that sum is intractable. Instead we make a series of variational
approximations: first we replace the sum with a maximization, then perform
iterated conditional modes, alternating between maximization of the conditional
probability of $\aligns$ and $\theta$.
We begin by initializing $\theta$ randomly.

As noted in the preceding section, the variable $b$ does not appear in these
equations. Conditioned on $\aligns$, the sum over structure-to-structure
$\psi(x,y) = \sum_b \psi(x,y,b)$ can be performed exactly using a simple dynamic
program which runs in time $\mathcal{O}(|x||y|)$ (assuming out-degree bounded by
a constant, and with $|x|$ and $|y|$ the number of words and graph nodes
respectively). This is \autoref{alg:dp}.



\begin{algorithm}[t]
  \caption{Computing structure-to-structure alignments}
  \label{alg:dp}
  \begin{algorithmic}
    \State $\txt^i$ are words in reverse topological order
    \State $\pth^j$ are grounding graph nodes (root last)
    \State \textit{chart} is an $m \times n$ array
    \For{$i = 1$ to $|\txt|$}
      \For{$j = 1$ to $|\pth|$}
        \State \textit{score} $\gets \exp \big\{\theta^\top \phi(\txt^i,
        y^j)\big\}$
        \For{$(k, l) \in d(i, j)$}
          
        \State $s \gets \sum_{l \in c(j)} \Big[ \exp \big\{\theta^\top \phi(\txt^{ik},
          y^{jl})\big\}$
        \State $\qquad\qquad\qquad \cdot\ \textit{chart}[k,l] \Big]$
        \State $\textit{score} \gets \textit{score} \cdot s$
        \EndFor
        \State $\textit{chart}[i,j] \gets score$
      \EndFor
    \EndFor
    \State \Return $\textit{chart}[n,m]$
  \end{algorithmic}
\end{algorithm}


In our experiments, $\theta$
is optimized using L-BFGS \cite{Liu89LBFGS}.
Calculation of the gradient with respect to $\theta$ requires
computation of a normalizing constant involving the sum over $p(\texts, \paths',
\aligns)$ for \emph{all} $\paths'$. While in principle the normalizing constant
can be computed using the forward algorithm, in practice the state spaces under
consideration are so large that even this is intractable. Thus we make an
additional approximation, constructing a set $\tilde{Y}$ of alternative actions
and taking \\
\scalebox{0.95}{\parbox{\columnwidth}{
\[ p(\paths, \aligns | \texts) \approx \sum_{j=1}^n \frac{
  \scriptstyle
  \exp\big\{ \psi(y_j) + \sum_{i=1}^m \indicate[a_i = j]\psi(x_i, y_i) \big\}
}{
  \scriptstyle
  \sum_{\tilde{y} \in \tilde{Y}} \exp\big\{ \psi(\tilde{y}) + \sum_{i=1}^m
  \indicate[a_i = j]\psi(x_i, \tilde{y})
\big\}
}
 \]
 }}\\
$\tilde{Y}$ is constructed by sampling
alternative actions from the environment model.  Meanwhile,
maximization of $\aligns$
can be performed exactly using the Viterbi algorithm,
without computation of normalizers.

Inference at test time involves a slightly different pair of optimization
problems. We again perform iterated conditional modes, here on the alignments
$\aligns$ and the unknown output path $\paths$. Maximization of $\aligns$ is
accomplished with the Viterbi algorithm, exactly as before; maximization of
$\paths$ also uses the Viterbi algorithm, or a beam search when this is
computationally infeasible. If bounds on path length are known,
it is straightforward to adapt these dynamic programs to efficiently consider
paths of all lengths.

\section{Evaluation}

As one of the main advantages of this approach is its generality, we evaluate on
several different benchmark tasks for instruction following. These exhibit great
diversity in both environment structure and language use. We compare our full
system to recent state-of-the-art approaches to each task. In the introduction,
we highlighted two core aspects of our approach to semantics: compositionality
(by way of grounding graphs and structure-to-structure alignments) and planning
(by way of inference with lookahead and sequence-to-sequence alignments).  To
evaluate these, we additionally present a pair of ablation experiments: \emph{no
grounding graphs} (an agent with an unstructured representation of environment
state), and \emph{no planning} (a reflex agent with no lookahead).

\paragraph{Map reading}

Our first application is the map navigation task established by \newcite{Vogel10SARSA}, based on
data collected for a psychological experiment by \newcite{Anderson91MapTask}
(\autoref{fig:tasks:hcrc}). Each training datum consists of a map with a
designated starting position, and a collection of landmarks, each labeled with a
spatial coordinate and a string name. Names are not always unique, and landmarks
in the test set are never observed during training. This map is accompanied by a
set of instructions specifying a path from the starting position to some
(unlabeled) destination point.
These instruction sets are informal
and redundant, involving as many as a hundred utterances.
They are transcribed from spoken text, so grammatical errors, disfluencies,
etc.\ are common.
This is a prime example of a domain that does not lend itself to logical
representation---grammars may be too rigid, and previously-unseen
landmarks and real-valued positions are handled more easily with feature
machinery than predicate logic.

\begin{table}
  \centering
  \small
  \begin{tabular}{lccc}
    \toprule
    & P & R & F$_1$ \\
    \midrule
    \newcite{Vogel10SARSA} & 0.46 & 0.51 & \bf 0.48 \\
    \newcite{Andreas14Paths} & 0.43 & 0.51 & 0.45 \\
    \whdashline
    Model [no planning] & 0.44 & 0.46 & 0.45 \\
    Model [no grounding graphs] & 0.52 & 0.52 & 0.52 \\
    Model [full] & 0.51 & 0.60 & \bf 0.55 \\
    \bottomrule
  \end{tabular}
  \caption{
    Evaluation results for the map-reading task. P is precision, R is recall and
    F$_1$ is F-measure. Scores are calculated with respect to transitions
    between landmarks appearing in the reference path (for details see
    \newcite{Vogel10SARSA}). We use the same train / test split. Some variant of
    our model achieves the best published results on all three metrics.
  }
  \label{tab:vogel-results}
\end{table}

The map task was previously studied by \newcite{Vogel10SARSA}, who implemented
{\sc sarsa} with a simple set of features. By combining these features with our
alignment model and search procedure, we achieve state-of-the-art results on
this task by a substantial margin (\autoref{tab:vogel-results}).


\begin{table}
  \centering
  \small
  \begin{tabular}{lS[table-format=1.2]}
    \toprule
    Feature & {Weight} \\
    \midrule
    {\tt\small word=top $\land$ side=North} & 1.31 \\
    {\tt\small word=top $\land$ side=South} & 0.61 \\
    {\tt\small word=top $\land$ side=East} & -0.93 \\
    \whdashline
    {\tt\small dist=0} & 4.51 \\
    {\tt\small dist=1} & 2.78 \\
    {\tt\small dist=4} & 1.54 \\
    \bottomrule
  \end{tabular}
  \caption{
    Learned feature values. The model learns that the word {\it top} often
    instructs the navigator to position itself above a landmark, occasionally to
    position itself below a landmark, but rarely to the side.
    The bottom portion of the table shows learned \emph{text-independent}
    constraints: given a choice, near destinations are preferred to far
    ones (so shorter paths are preferred overall).
  }
  \label{tab:vogel-feats}
\end{table}

Some learned feature values are shown in \autoref{tab:vogel-feats}. The model
correctly infers cardinal directions (the example shows the preferred side of a
destination landmark modified by the word {\em top}). Like Vogel et al., we
see support for both allocentric references ({\it you are on top of the
hill}) and egocentric references ({\it the hill is on top of you}).
We can also see pragmatics at work: the model learns useful text-independent
constraints---in this case, that near destinations should be preferred to far
ones.

\paragraph{Maze navigation}

The next application we consider is the maze navigation task of
\newcite{MacMahon06SAIL} (\autoref{fig:tasks:sail}). Here, a virtual agent is
situated in a maze (whose hallways are distinguished with various wallpapers,
carpets, and the presence of a small set of standard objects), and again given
instructions for getting from one point to another. This task has been the
subject of focused attention in semantic parsing for several years, resulting
in a variety of sophisticated approaches.

Despite superficial similarity to the previous navigation task, the language and
plans required for this task are quite different. The proportion of instructions
to actions is much higher (so redundancy much lower), and the interpretation of
language is highly compositional.

As can be seen in \autoref{tab:chen-results}, we outperform a number of systems
purpose-built for this navigation task. We also outperform both variants of our
system, most conspicuously the variant without grounding graphs. This highlights
the importance of compositional structure.  Recent work by
\newcite{Kim13Reranker} and \newcite{Artzi14Compact} has achieved better
results; these systems make use of techniques and resources (respectively,
discriminative reranking and a seed lexicon of hand-annotated logical forms)
that are largely orthogonal to the ones used here, and might be applied to
improve our own results as well. 

\begin{table}
  \centering
  \small
  \begin{tabular}{l*{1}{S[table-format=2.1]}}
    \toprule
    & {Success (\%)} \\
    \midrule
    \newcite{Kim12Instructions} & 57.2 \\
    \newcite{Chen12Online} & {\bf 57.3} \\
    \whdashline
    Model [no planning] & 58.9 \\
    Model [no grounding graphs] & 51.7 \\
    Model [full] & {\bf 59.6} \\
    \whdashline
    \newcite{Kim13Reranker} [reranked] & 62.8 \\
    \newcite{Artzi14Compact} [semi-supervised] & {\bf 65.3} \\
    \bottomrule
  \end{tabular}
  \caption{Evaluation results for the maze navigation task. ``Success'' shows the
  percentage of actions resulting in a correct position and orientation after
  observing a single instruction. We use the leave-one-map-out evaluation
  employed by previous work.\footnotemark\ 
  All systems are trained on full action sequences.  Our model outperforms
  several task-specific baselines, as well as a baseline with path structure but
  no action structure.}
  \label{tab:chen-results}
\end{table}

\footnotetext{We specifically targeted the single-sentence version of this
evaluation, as an alternative full-sequence evaluation does not align precisely
with our data condition.}

\paragraph{Puzzle solving}

The last task we consider is the Crossblock task studied by
\newcite{Branavan09PG} (\autoref{fig:tasks:crossblock}). Here, again, natural
language is used to specify a sequence of actions, in this case the solution to
a simple game. The environment is simple enough to be captured with a flat
feature representation, so there is no distinction between the full model and
the variant without grounding graphs.

Unlike the other tasks we consider, Crossblock is distinguished by a
challenging associated search problem. Here it is nontrivial to find \emph{any}
sequence that eliminates all the blocks (the goal of the puzzle). Thus this
example allows us measure the effectiveness of our search procedure.

Results are shown in \autoref{tab:branavan-results}. As can be seen, our model
achieves state-of-the-art performance on this task when attempting to match
the human-specified plan exactly. If we are purely concerned with task
completion (i.e. solving the puzzle, perhaps not with the exact set of moves
specified in the instructions) we can measure this directly. Here, too, we
substantially outperform a no-text baseline. 
Thus it can be seen that text induces a useful heuristic, allowing the model to
solve a considerable fraction of problem instances not solved by na\"ive beam
search.

The problem of inducing planning heuristics from side information like text is
an important one in its own right, and future work might focus
specifically on coupling our system with a more sophisticated planner. Even at
present, the results in this section demonstrate the importance of
lookahead and high-level reasoning in instruction following.


\begin{table}
  \centering
  \small
  \begin{tabular}{l*{2}{S[table-format=2]}}
    \toprule
    & {Match (\%)} & {Success (\%)} \\
    \midrule
    No text & 54 & 78 \\
    Branavan '09 & {\bf 63} & {--} \\
    \whdashline
    Model [no planning] & 64 & 66 \\
    Model [full] & {\bf 70} & {\bf 86} \\
    \bottomrule
  \end{tabular}
  \caption{Results for the puzzle solving task. ``Match'' shows the percentage
  of predicted action sequences that exactly match the annotation.
  ``Success'' shows the percentage of predicted action sequences that result in
  a winning game configuration, regardless of the action sequence performed. 
  Following \newcite{Branavan09PG}, we average across five random train / test
folds.  Our model achieves state-of-the-art results on this task.}
  \label{tab:branavan-results}
\end{table}

%

\section{Conclusion}

We have described a new alignment-based compositional model for following
sequences of natural language instructions, and demonstrated the effectiveness
of this model on a variety of tasks.  A fully general solution to the problem of
contextual interpretation must address a wide range of well-studied problems,
but the work we have described here provides modular interfaces for the study of
a number of fundamental linguistic issues from a machine learning perspective.
These include:

\paragraph{Pragmatics}
How do we respond to presupposition failures, and choose among possible
interpretations of an instruction disambiguated only by context? The mechanism
provided by the sequence-prediction architecture we have described provides a
simple answer to this question, and our experimental results demonstrate that
the learned pragmatics aid interpretation of instructions in a number of
concrete ways: ambiguous references are resolved by proximity in the map reading
task, missing steps are inferred from an environment model in the maze
navigation task, and vague hints are turned into real plans by knowledge of the
rules in Crossblock.  A more comprehensive solution might explicitly describe
the process by which instruction-givers' own beliefs (expressed as distributions
over sequences) give rise to instructions.

\paragraph{Compositional semantics}
The graph alignment model of semantics presented here is an expressive and
computationally efficient generalization of classical logical techniques to
accommodate environments like the map task, or those explored in our previous
work \cite{Andreas14Paths}. More broadly, our model provides a compositional
approach to semantics that does not require an explicit formal language for
encoding sentence meaning. Future work might extend this approach to tasks like
question answering, where logic-based approaches have been successful.

\paragraph{}
Our primary goal in this paper has been to explore methods for integrating
compositional semantics and the pragmatic context provided by sequential
structures. While there is a great deal of work left to do, we find it
encouraging that this general approach results in substantial gains across
multiple tasks and contexts.

\section*{Acknowledgments}

The authors would like to thank S.R.K.\ Branavan for assistance with the
Crossblock evaluation. The first author is supported by a National Science
Foundation Graduate Fellowship.

\bibliography{jacob}
\bibliographystyle{acl2012}

\end{document}